# How to achieve various gait patterns from single nominal

**Miomir Vukobratović[1)], Dejan Andrić[2)], Branislav Borovac[2)]**
[1)] Mihajlo Pupin Institute, Volgina 15 Str., 11000-Belgrade, Serbia and Montenegro
e-mail: vuk@robot.imp.bg.ac.yu
[2)] Faculty of Technical Sciences, University of Novi Sad, D. Obradovića Sq. 6, 21000-Novi Sad,
Serbia and Montenegro, e-mail:{zoro, borovac@uns.ns.ac.yu}

*Abstract:* In this paper is presented an approach to achieving on-line modification of nominal biped gait without recomputing entire dynamics when steady motion is performed. Straight, dynamically balanced walk was used as a nominal gait, and applied modifications were speed-up and slow-down walk and turning left and right. It is shown that the disturbances caused by these modifications jeopardize dynamic stability, but they can be simply compensated to enable walk continuation.

## 1. Introduction

Humanoid robotics has been in focus of scientific community for decades. Among other topics biped locomotion is one of the most important because it is expected that robots will share the same space with humans.

Irrespective of their structure and number of degrees of freedom (DOF) involved, the basic characteristics of all biped locomotion systems are: a) the presence of *unpowered* DOFs formed by the contact of the foot with the ground surface, b) gait repeatability, and c) regular interchangeability of the single- and double-support phases. During the walk, two different situations arise in sequence: the statically stable double-support phase, in which the mechanism is supported on booth feet simultaneously, and statically unstable single-support phase, when only one foot of the mechanism is in contact with the ground while the other is being transferred from the back to front position. Thus, the locomotion mechanism changes its structure during the single walking cycle from an open to a closed kinematic chain.

All of the biped mechanism joints are powered and directly controllable except for the contact of the foot and the ground (it can be considered as an additional DOF), which is the only point at which the mechanism interacts with the environment. This contact is essential for the walk realization because the mechanism's position with respect to the environment depends on the relative position of the foot with respect to the ground.

The foot cannot be controlled directly but in an indirect way – by ensuring appropriate dynamics of the mechanism above the foot. Thus, the overall indicator of the mechanism behavior is the point where the influence of all the forces acting on the mechanism can be replaced by one single force. This point was termed *Zero-Moment Point* (*ZMP*) (Vukobratović M. & Juričić D. 1968, Vukobratović M. & Juričić D. 1969, Vukobratović M. & Stepanen-ko Yu., 1972). Recognition of the significance and role of ZMP in biped artificial walk was a turning point in gait planning and control.

In the gait synthesis by semi-inverse method (Vukobratović M. & Juričić D. 1968, Vukobratović M. & Juričić D. 1969, Vukobratović M. & Stepanen-ko Yu., 1972) legs trajectories are predefined while the trunk motion is determined so as to ensure dynamic equilibrium of the system as a whole (the ZMP position is within the desired area under the supporting foot). Such motion is called nominal motion.

In contemporary practice, the aim is to have the possibility of synthesizing all gait patterns (under a gait pattern we understand each gait with specifically defined parameters), and each synthesized gait is called nominal. However, such approach requires the synthesis of a huge number of different patterns, and a question arises whether one can envisage all possible cases that may be of interest, so that for each situation a most suitable one could be selected and executed. In addition, during the gait realization, a situation can often arise when small modifications of the basic regime are needed to adapt to the momentary requirements imposed on the locomotion system. We believe that, at least in some basic cases, it is not necessary to perform a new nominal synthesis but is possible to do on-line modification of an existing nominal (Vukobratović *et al.* 2003, (Vukobratović *et al.* 2004).

**2. Mechanism structure and its nominal motion**

Our idea was tested on the example of a locomotion mechanism (Fig. 1) having 20 DOFs, with all joints being rotational (Vukobratović *et al.* 1990). Between two joints is placed one link. A joint having more DOFs is modeled as a set of simple rotational joints, each having only one DOF. In the case of a multi-DOF joint modeling the links between single-DOF joints are considered as being of zero length, mass and moments of inertia. In Fig. 1, these "fictitious" links are denoted by dashed line (for example link 3 at the ankle joint, links 6 and 7 at the hip joint of the right leg, etc.). Accordingly, the ankle joint having 2 DOFs is represented by joints 3 and 4 (the joint axes are defined by the orts $e_3$ and $e_4$), hip joint of the right leg by joints 6, 7 and 8, trunk joint (waist) by joints 15 and 16. All other joints are modeled by simple rotational joints having one DOF.

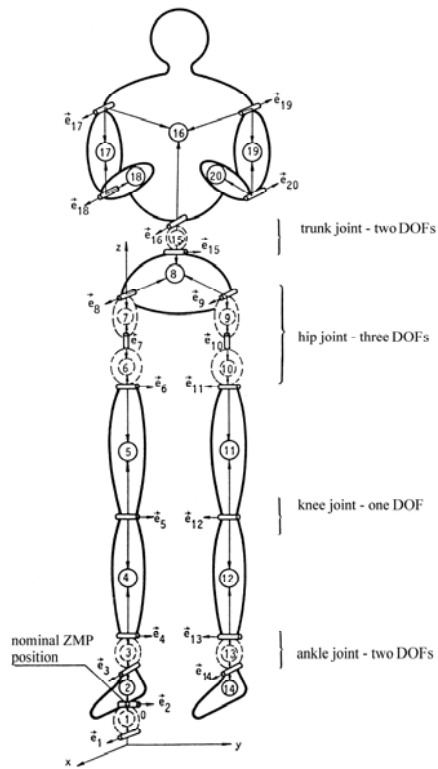

Fig. 1. Scheme of locomotion mechanism

The possibility of the mechanism's overturning about the foot edge is modeled by two "fictitious" joints in the directions of the axes *x* and *y* to detect if the mechanism as a whole will rotate in the *y-z* or *x-z* plane. Each mechanism joint is powered by a DC motor except for joints 1 and 2 under the foot representing unpowered DOFs. All parameters of the mechanism can be found in [4].

The mechanism motion is synthesized using semi-inverse method in such a way that dynamic equilibrium is permanently ensured during the walk (Vukobratović *et al.* 1990). Legs motion pattern was obtained by recording the performance of a human subject, and then, the trunk motion was synthesized in such a way to ensure that ground reaction force under the foot is in a certain predefined position (in our example this is the point O of the coordinate system $O_{xyz}$), ensuring simultaneously that horizontal components of the ground reaction moment are equal to zero, i.e. $M_x=M_y=0$. As already said, this point is known as ZMP. Each change of the dynamics above the supporting foot causes displacement of the ZMP out of its nominal position. Walk synthesis has been carried out for one half-step of single-support phase only. The mechanism walks in such a way that while one leg is in support phase the other one is transferred from the back to front position. When it reaches the position just above the ground floor the support foot is changed and the walk continues. Footprints of the mechanism nominal motion (ZMP nominal positions) are presented in Fig. 2.

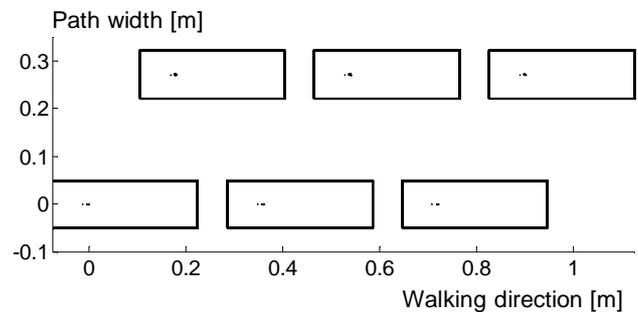

Fig. 2. Mechanism footprints and ZMP position during nominal motion

If during the walk the ZMP position comes out of the support polygon, this means that the mechanism is not performing dynamically balanced gait any more, and it collapses by rotating about the foot edge. ZMP position, being very sensitive to the changes of dynamics of the mechanism above the foot, is the best overall indicator of the mechanism dynamic equi-librium. In Fig. 3. is shown a stick diagram of the mechanism performing nominal motion.

**3. Modification of the nominal motion**

*3.1 Turning*
The nominal motion modification to be considered first is the change of walk direction. Namely, very often there

100

is a need to switch from a rectilinear motion to the right/left direction without interrupting the walk, so that it would be highly desirable to achieve such a gait by modifying nominal rectilinear motion.

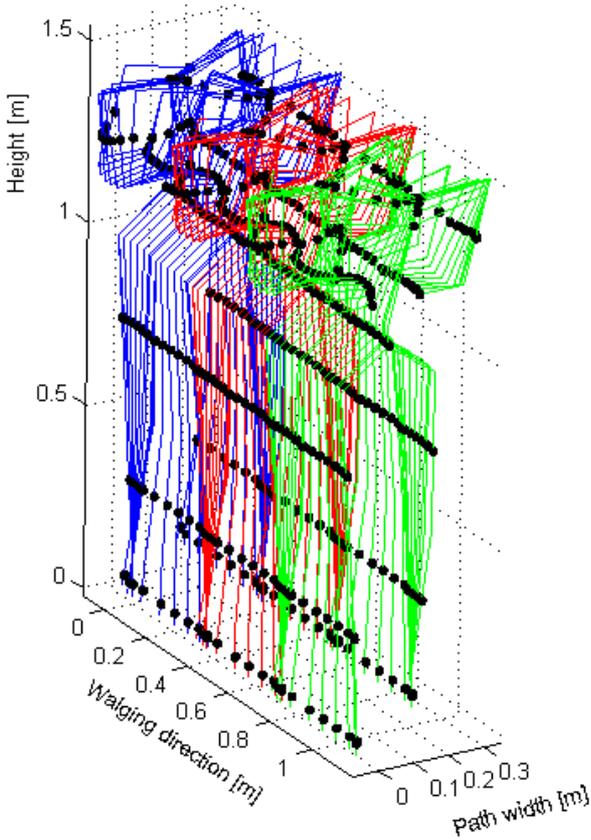

Fig. 3. Stick diagram of the mechanism performing nominal motion

First thing to be considered is how to perform turning. The angle α formed between the foot at the end of a single-support phase and the foot that up to that moment was in contact with the ground will be adopted as the measure of the extent of turning.

A smaller α corresponds to a "milder" and a larger to a "sharper" turning. Besides, care should be exercised as to the relative position of the mechanism's feet (legs) with respect to the "body". In the case of the mechanism used (see Fig. 2) the turning will be realized by rotation at joints 7 and 10, so that both joints are activated simultaneously.

A simplest algorithm would be to perform turning so that in each half-step the leg being in the swing phase rotates so that the foot that is making contact with the ground forms an angle α with respect to the foot that was previously in contact with the ground. However, this algorithm did not perform well since the "outer" and "inner" leg followed arcs of different lengths so that soon arose the situation in which the motion could not be continued at all. A solution to this problem would be to perform deflection in the first half-step, and in the second correction of the deviation due to the foot rotation, so that in the beginning of the next step the feet are not deflected with respect to the trunk.

The next thing to be considered is the way in which foot turning by the angle α will be realized – whether once or incrementally in the course of the whole half-step, so that at the end the foot deflects by the full value of the angle α. In the present work we chose gradual increase of the turning angle.

Let the desired value of foot deflection at the end of a half-step be defined by the angle α. If the number of integration intervals during a half-step is $n_{int}$ the value of deflection angle to be realized in one integration interval is $\Delta\alpha = \alpha/n_{int}$. This value is halved and the value ($\Delta\alpha/2$) is added to each of joints 7 and 10. Let us explain this on the example of turning to the left, as shown in Fig. 4. During the first half-step, the right leg is in the support and the left leg in the swing phase. The angles are added in the following way:

$$\begin{aligned} \textbf{\textit{angle}}\,7 &= \textbf{\textit{angle}}\,7_{nom} + \textbf{\textit{i}} \cdot (\Delta\alpha/2) \\ \textbf{\textit{angle}}\,10 &= \textbf{\textit{angle}}\,10_{nom} + \textbf{\textit{i}} \cdot (\Delta\alpha/2) \end{aligned} \quad (1)$$

where *i* denotes ordinal number of the integration interval while $\textbf{\textit{angle}}\,7_{nom}$ and $\textbf{\textit{angle}}\,10_{nom}$ are angle values for nominal motion. In the end of this phase, the left foot is deflected relative to the right foot by angle α (note that first and second footsteps are not parallel). In the second half-step, to compensate for the deviation of legs positions relative to the mechanism body the hip positions change in the following way:

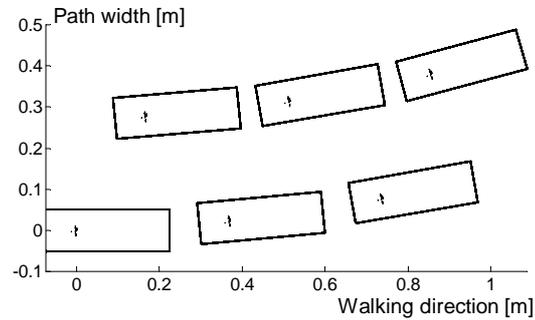

Fig. 4. Mechanism footprints when α is 5°. No compensation applied

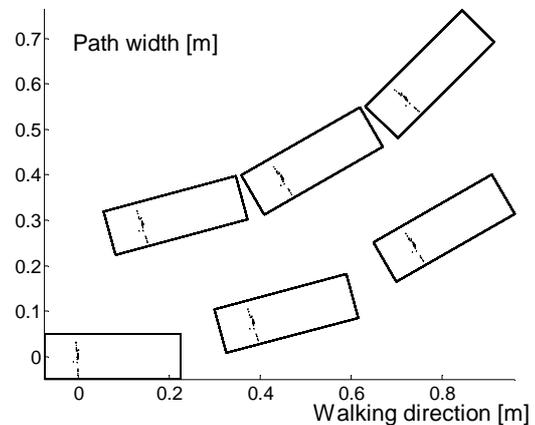

Fig. 5. Mechanism footprints when α is 15°. No compensation applied.

$$angle\ 7 = -(angle\ 7_{nom} + i \cdot (\Delta\alpha/2))$$
$$angle\ 10 = -(angle\ 10_{nom} + i \cdot (\Delta\alpha/2)) \quad (2)$$

In the end of the second half-step the right foot becomes parallel to the left foot. After that, the procedure continues and the left foot angle relative to its previous position increases again, the right foot compensates for this deviation, and so on.

In the following figures are given examples of the algorithm application. In Figs. 4, 5 and 6 are presented footprints for the angle $\alpha$ of $5^\circ$, $15^\circ$ and $25^\circ$, respectively, to illustrate how the change of walking direction affects dynamic equilibrium of nominal motion. From Fig. 4 it is clear that ZMP is close to its nominal position (far enough from the foot edge), and no compensation is needed. However, as is evident from Figs. 5 and 6 when angle $\alpha$ increases, ZMP deviation from its nominal position increases too. In Fig. 5 it is very close to the foot edge, while in Fig. 6 it goes out of the support polygon and correction of ZMP position is necessary. To compensate for ZMP deviation we propose to add a permanent gravitational moment which can be produced by body inclination. Now, a question arises which joint is most suitable for this operation. We investigated all possibilities (trunk, hip, and ankle) and fond that compensation by ankle gives the best results. In Fig. 7 are given footprints for the same case as shown in Fig. 6 ($\alpha = 25^\circ$) but with compensation at joint 3 of $3^\circ$. It can be seen that ZMP position deviations are significantly reduced and that all ZMP positions are within the support polygon. Thus, by such compensation dynamic equilibrium is preserved and the walk continuation ensured.

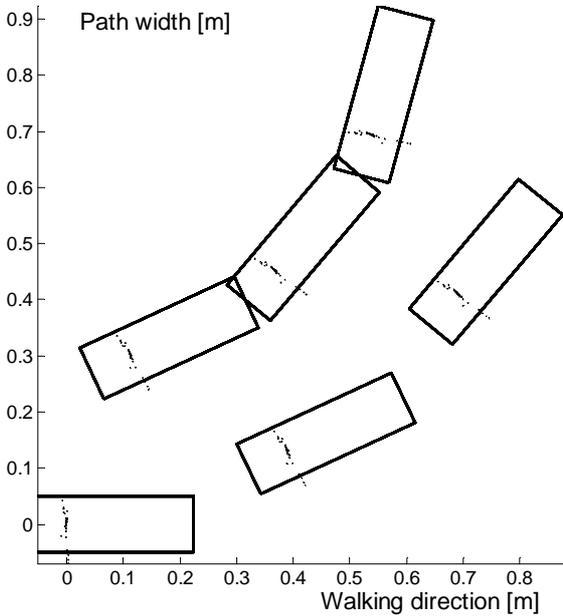

Fig. 6. Mechanism footprints when $\alpha$ is $25^\circ$. No compensation is applied.

*3.2 Walk speed-up and slow-down*

The next case we investigated was how to speed up and slow down the walk. Nominal motion was the same as before.

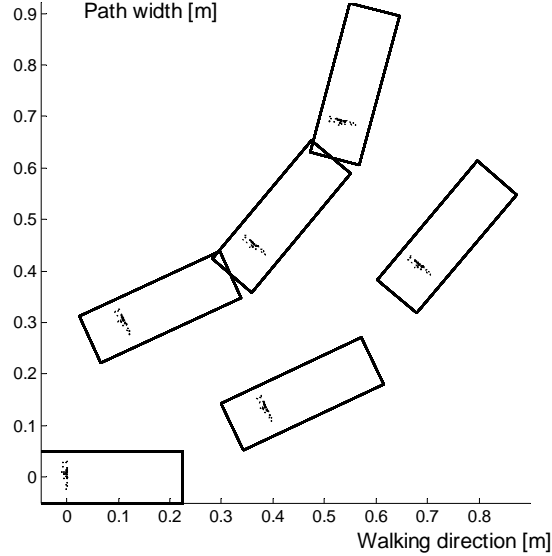

Fig. 7. Mechanism footprints when $\alpha$ is $25^\circ$. Compensation at the ankle joint (joint 3) $3^\circ$.

To change walking speed, moments applied at all joints have to be changed. To avoid recomputing complete dynamics we used the procedure proposed by (Hollerbach J. M. 1984) for changing robotic mechanism motion speed without recomputing completely its dynamics.

Let us remind that the mechanism dynamics is given in the following general form:

$$\tau = H(q)\ddot{q} + h(\dot{q}) + G \quad (3)$$

where $\tau$ is the vector of driving torques at joints, $q$, $\dot{q}$ and $\ddot{q}$ are the vectors of joint angles, velocities and accelerations, respectively. $H(q)$ and $h(\dot{q})$ are the inertial matrix and the vector that includes all velocity effects, while vector $G$ represents gravi-tational moments. Gravitational forces and their mo-ments do not depend on motion speed. Thus, in (3) we can separate the moments that are acceleration- and velocity-dependent from those which are not. If we denote by $\tau_n$ acceleration- and velocity-depen-dent torques of non-accelerated motion and by $\tau_a$ those of accelerated motion, then the

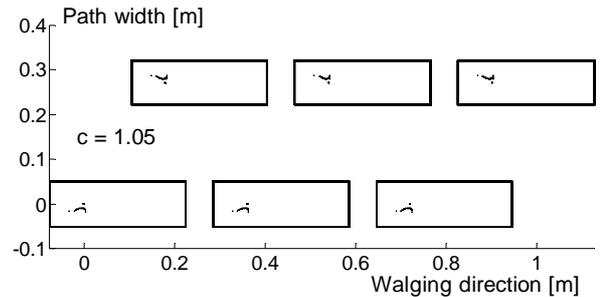

Fig. 8. Mechanism footprints of accelerated motion for $c$ =1.05. No compensation applied.

following relation (Hollerbach J. M. 1984) will hold:
$$\tau_a(t) = c^2 \cdot \tau_n(ct) \qquad (4)$$
whereby for $c > 1$ the motion is accelerated, and for $c < 1$ decelerated.

We applied this method to investigate how the acceleration/deceleration of nominal (dynamically bala-nced) walk would affect its dynamic equilibrium. In Figs. 8-12 are shown the cases of walk speeding up for various values of c.

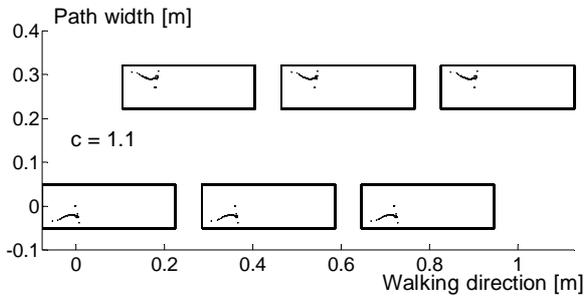

Fig. 9. Mechanism footprints of accelerated motion for $c$ =1.1. No compensation applied.

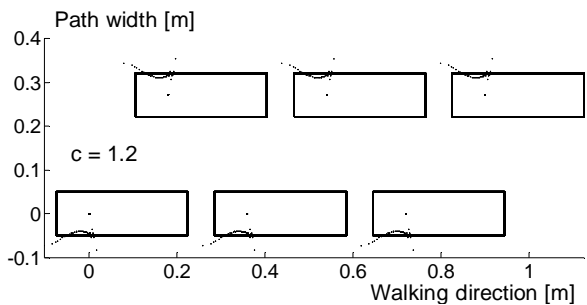

Fig. 10. Mechanism footprints of accelerated motion for $c$ =1.2. No compensation applied.

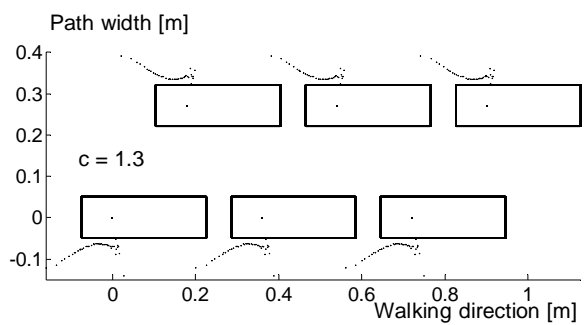

Fig. 11. Mechanism footprints of accelerated motion for $c$ =1.3. No compensation applied.

The presented figures illustrate well the importance of the altered mechanism dynamics on the ZMP position. For $c > 1.1$ (Figs. 10 – 12) it is clear that ZMP is out of the support polygon and that mechanism dynamic equilibrium is not preserved. To make possible continuation of the walk it is necessary to correct the ZMP position.

To achieve this we applied the same strategy as before –

inclination of the body. In Figs. 13-17 are presented the correction effects when compensation was performed by joints 3 and 4 (ankle). Joint 3 compensated for deviations in the sagittal and joint 4 in the frontal plane.

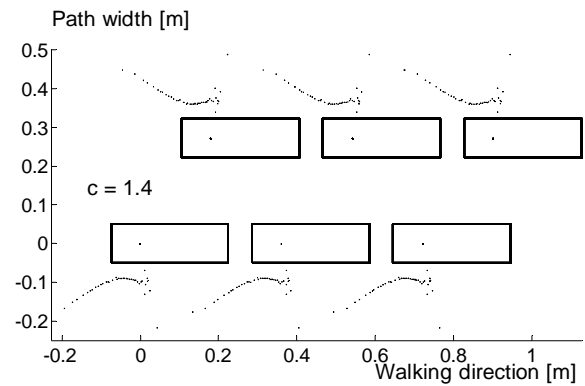

Fig. 12. Mechanism footprints of accelerated motion for $c$ =1.4. No compensation applied.

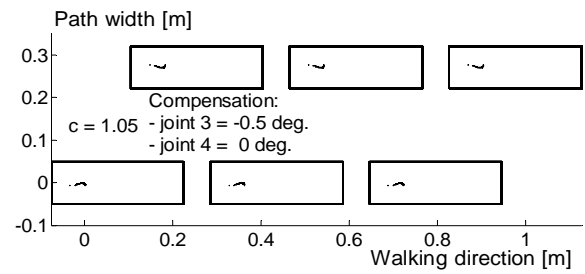

Fig. 13. Mechanism footprints of accelerated motion for $c$ =1.05. Compensation applied at joint 3 –0.5°, at joint 4 no compensation.

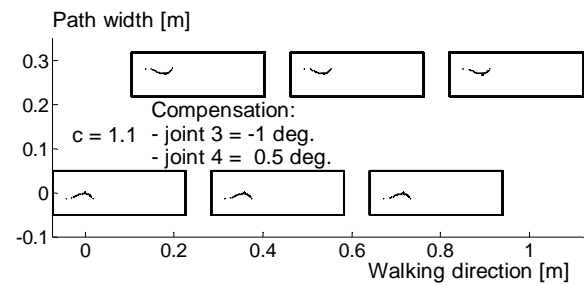

Fig. 14. Mechanism footprints of accelerated motion for $c$ = 1.1. Compensation applied at joint 3 –2°, at joint 4 0.5°.

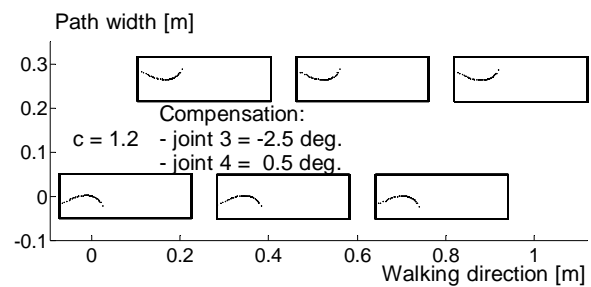

Fig. 15. Mechanism footprints of accelerated motion for $c$ =1.2. Compensation applied at joint 3 –2.5°, at joint 4 0.5°.

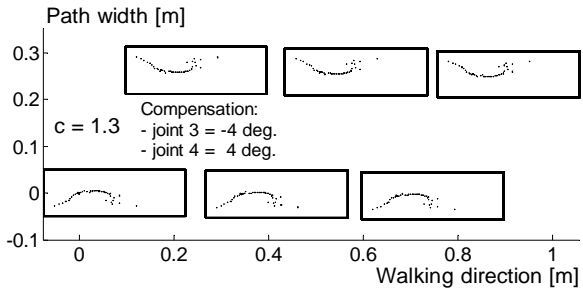

Fig. 16. Mechanism footprints of accelerated motion for $c$ =1.3. Compensation applied at joint 3 –4°, at joint 4 4°.

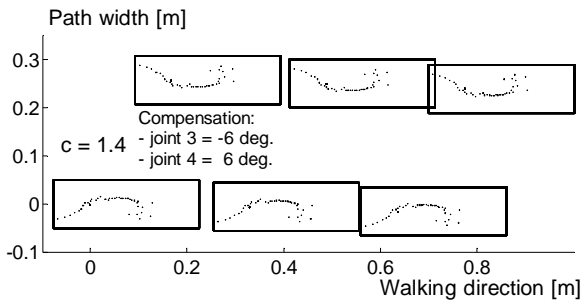

Fig. 17. Mechanism footprints of accelerated motion for $c$ =1.4. Compensation applied at joint 3 –6°, at joint 4 6°.

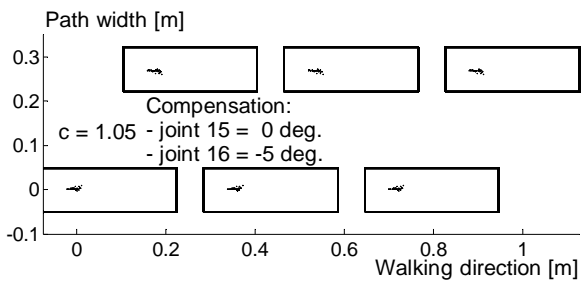

Fig. 18. Mechanism footprints of accelerated motion for $c$ =1.05. Compensation applied at joint 15 0°, at joint 16 -5°.

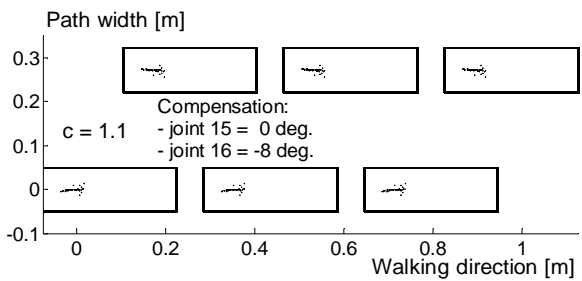

Fig. 19. Mechanism footprints of accelerated motion for $c$ =1.1. Compensation applied at joint 15 0°, at joint 16 -8°.

The presented examples show that the corrections made brought ZMP back to the support polygons. Figs. 18-22 illustrate the situations when the ZMP deviation was compensated by trunk joints (joints 15 and 16). It is evident that much larger inclination angles are needed for compensation. Such large compensation inclinations may jeopardize anthro-pomorphic pattern of the walk. Besides, for $c$ =1.3 and 1.4 the ZMP positions were not brought back to the support polygon.

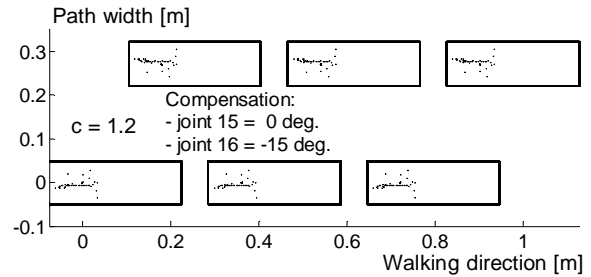

Fig. 20. Mechanism footprints of accelerated motion for $c$ =1.2. Compensation applied at joint 15 0°, at joint 16 -15°.

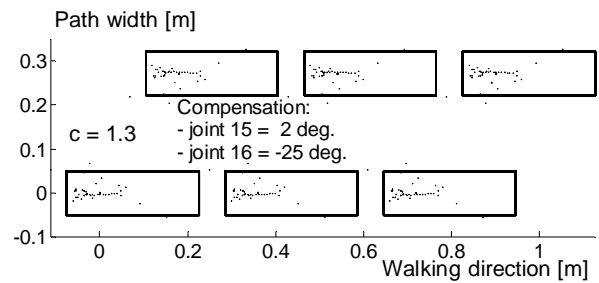

Fig. 21. Mechanism footprints of accelerated motion for $c$ =1.3. Compensation applied at joint 15 2°, at joint 16 -25°.

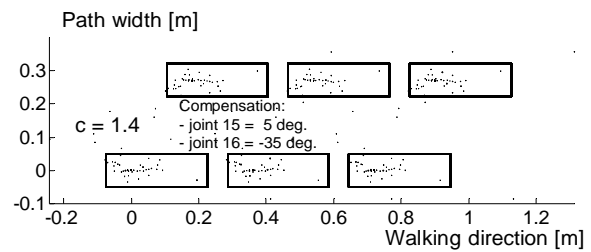

Fig. 22. Mechanism footprints of accelerated motion for $c$ =1.4. Compensation applied at joint 15 5°, at joint 16 -35°.

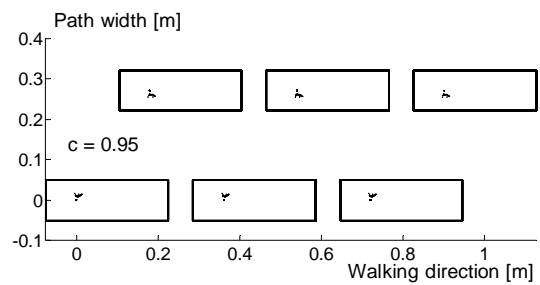

Fig. 23. Mechanism footprints of decelerated motion for $c$ =0.95. No compensation applied.

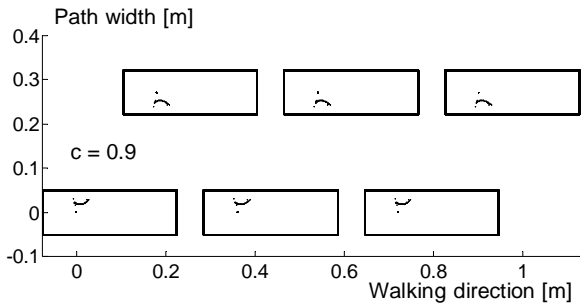

Fig. 24. Mechanism footprints of decelerated motion for *c* =0.9. No compensation applied.

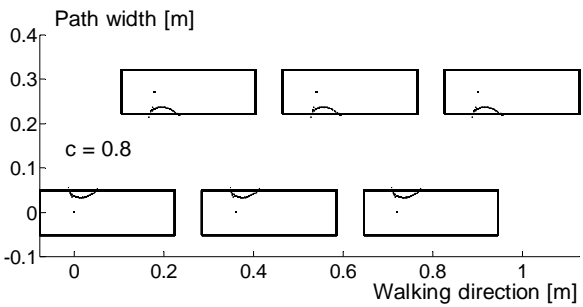

Fig. 25. Mechanism footprints of decelerated motion for *c* =0.8. No compensation applied.

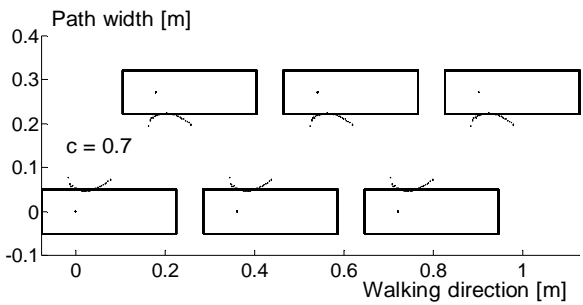

Fig. 26. Mechanism footprints of decelerated motion for *c* =0.7. No compensation applied.

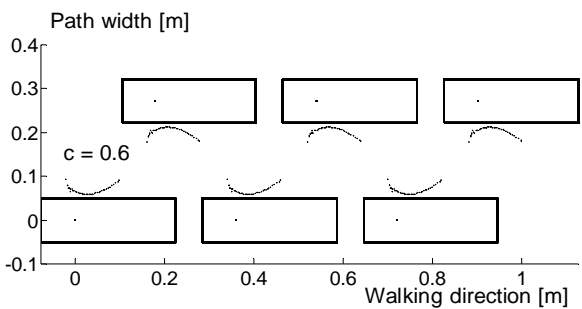

Fig. 27. Mechanism footprints of decelerated motion for *c* =0.6. No compensation applied.

Dispersion of ZMP positions are significant, too. Hence it can be concluded that the ankle is more suitable for this purpose than the trunk.
In Figs. 23-27 are presented examples of decelarated motion (c = 0.95 – 0.5). It is evident that spatial dissipation of the ZMP trajectory is lower. This can be explained by the diminished overall influence of inertial forces, so that the ZMP trajectory converges to the trajectory of the system's gravity center, which is a sinusoide lying symmetrically between the mechanism's footprints.

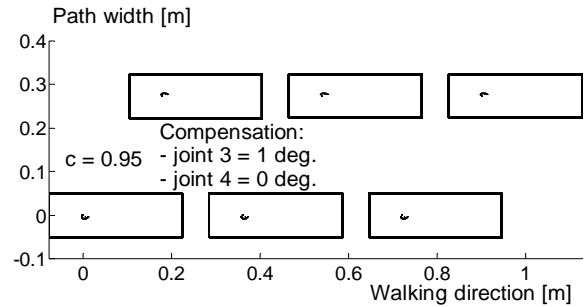

Fig. 28. Mechanism footprints of decelerated motion for *c* =0.95. Compensation applied at joint 3  1°, at joint 4 no compensation.

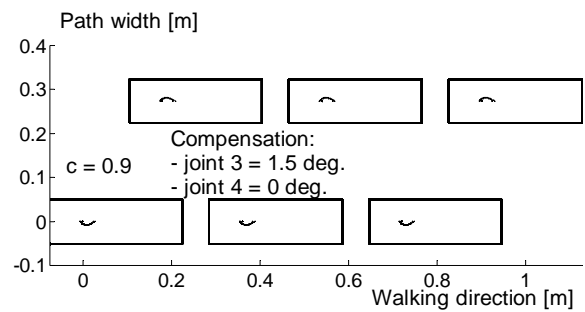

Fig. 29. Mechanism footprints of decelerated motion for *c* = 0.9. Compensation applied at joint 3  1.5°, at joint 4 no compensation.

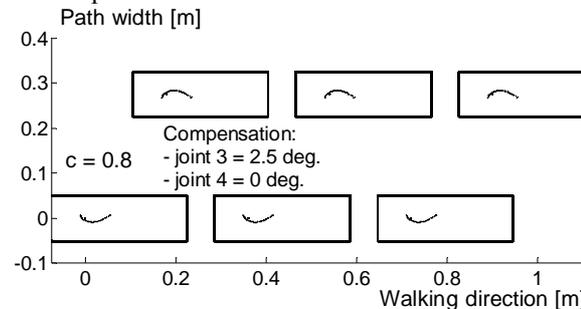

Fig. 30. Mechanism footprints of decelerated motion for *c* =0.8. Compensation applied at joint 3  2.5°, at joint 4 no compensation.

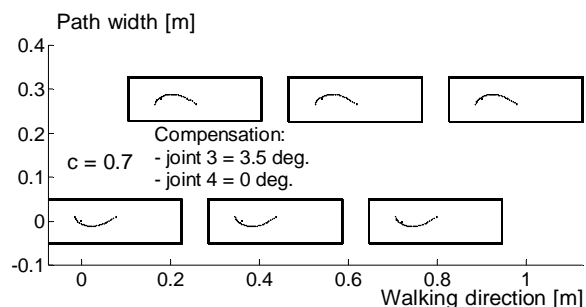

Fig. 31. Mechanism footprints of decelerated motion for *c* =0.7. Compensation applied at joint 3  3.5°, at joint 4 no compensation.

To compensate for ZMP deviations we applied again the same strategy, i.e. body inclination. as can be seen, both the compensation by the ankle joint (figs 28-32) and by the trunk (figs 33-37) was successful.

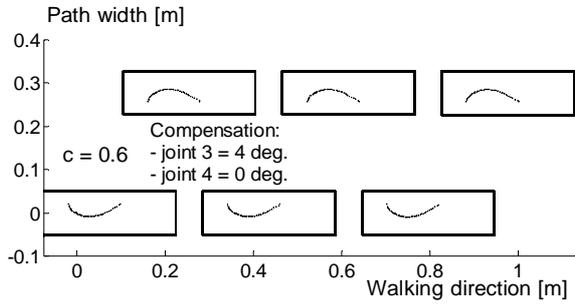

Fig. 32. Mechanism footprints of decelerated motion for $c = 0.6$. Compensation applied at joint 3  4°, at joint 4 no compensation.

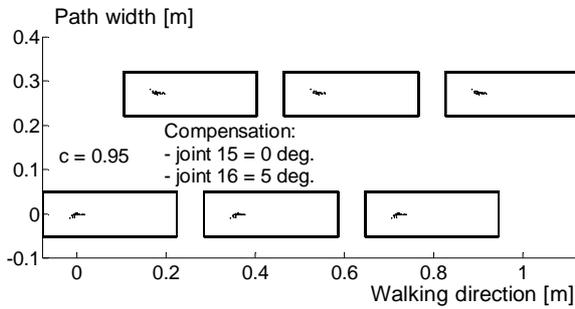

Fig. 33. Mechanism footprints of decelerated motion for $c = 0.95$. Compensation applied at joint 15  0°, at joint 16  5°.

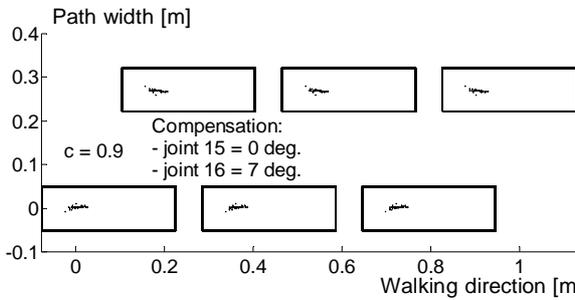

Fig. 34. Mechanism footprints of decelerated motion for $c = 0.9$. Compensation applied at joint 15  0°, at joint 16  7°.

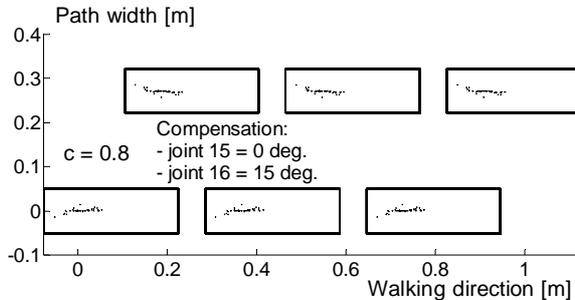

Fig. 35. Mechanism footprints of decelerated motion for $c = 0.8$. Compensation applied at joint 15  0°, at joint 16  15°.

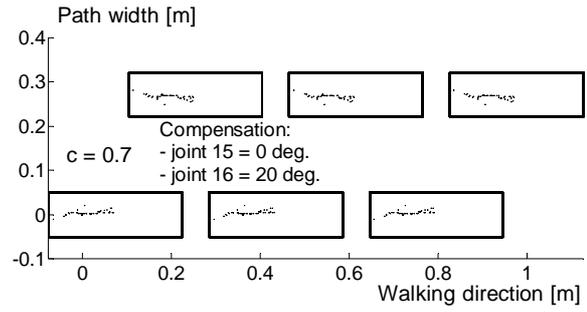

Fig. 36. Mechanism footprints of decelerated motion for $c = 0.7$. Compensation applied at joint 15  0°, at joint 16  20°.

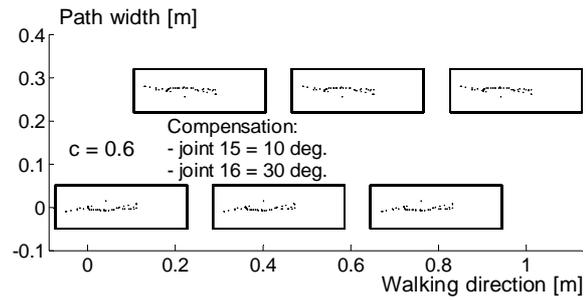

Fig. 37. Mechanism footprints of decelerated motion for $c = 0.6$. Compensation applied at joint 15  10°, at joint 16  30°.

### 3.3 Step extention

The next basic modification of the nominal walk is change of step length. In this paper step extention will be considered, only. To extend step maximal angle between two legs for nominal step (in Fig. 38 denoted by $\beta_{nom}$) should be enlerged.

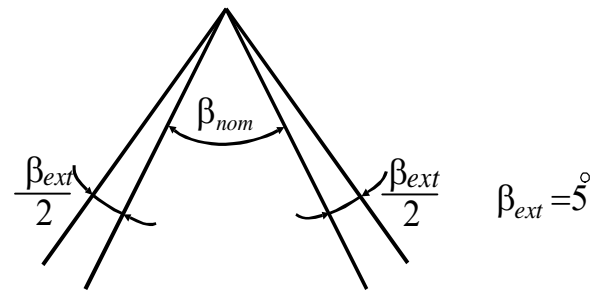

Fig. 38. Angles $\beta_{nom}$ and $\beta_{ext}$

Let us denote total value of angle added as $\beta_{ext}$. Total amount of $\beta_{ext}$ is divided between two hip joints (joint 6 for right, joint 11 for left leg) and, accordingly per each leg is added $\beta_{ext}/2$. Thus, maximal angle between legs for extended step is $\beta = \beta_{nom} + \beta_{ext}$. As in case of turning, angle beta should be enlarged gradually. This means that in each sampling time actual value of angles 6 and 11 is enlarged by $\Delta\beta = \beta_{ext}/n_{int}$, where $n_{int}$ is number of integration intervals during a half-step:

$$angle\,6 = angle\,6_{nom} + i \cdot (\Delta\beta/2)$$
$$angle\,11 = angle\,11_{nom} + i \cdot (\Delta\beta/2) \qquad (5)$$

where, again, $i$ denotes ordinal number of the integration interval while $angle\,6_{nom}$ and $angle\,11_{nom}$ are angle values for nominal motion. Of course, to preserve position of the foot relative to the ground, ankle joints 4 and 13 should be corrected accordingly.

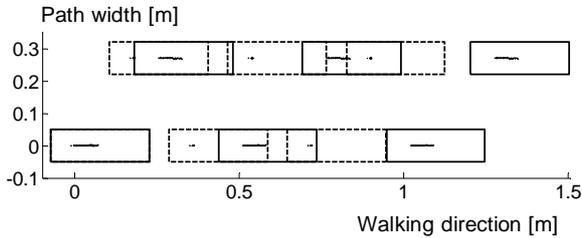

Fig. 39. Extended step

In Fig. 39 are given footsteps of such extended gait. Dashed line denotes footsteps of the nominal gait, solid line of enlarged stride. It can be noticed that the ZMP position deviates along the footprint axis and, actually, no correction is needed.

### 4. Combination of basic modifications of the nominal gait

*4.1. Step extention and spped-up for rectilinear and turning motion*

In Figs. 40-49 are given examples of the combina-tions of basic modifications for rectilinear motion: gait speed-up and step extension. In Figs. 40 and 42 are given examples of noncompensated and in Figs. 41 and 43 are given compensated motion. Because of space limitation only two examples of speed-up gait are presented, for $c$=1.2 and $c$=1.3.

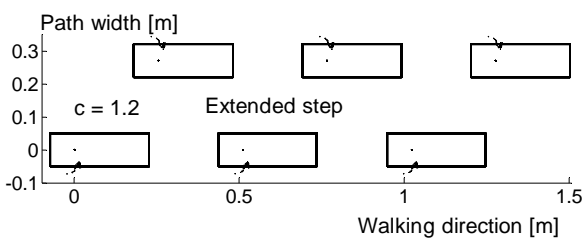

Fig. 40. Mechanism footprints of extended and accelerated motion for $c = 1.2$. No compensation applied.

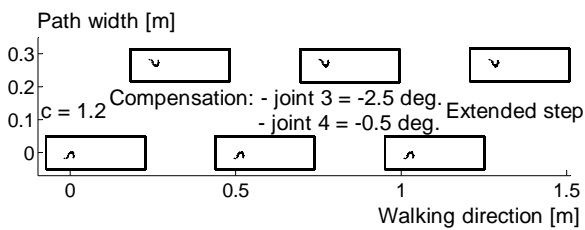

Fig. 41. Mechanism footprints of extended and accelerated motion for $c = 1.2$. Compensation applied at joint 3 $-2.5°$, at joint 4 $-0.5°$.

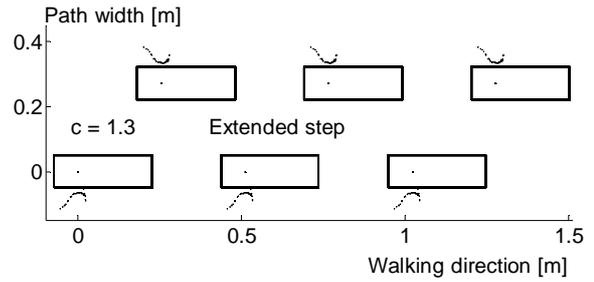

Fig. 42. Mechanism footprints of extended and accelerated motion for $c = 1.3$. No compensation applied.

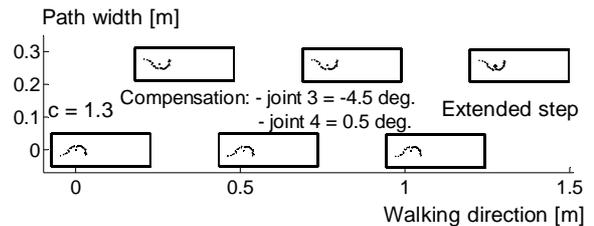

Fig. 43. Mechanism footprints of extended and accelerated motion for $c = 1.3$. Compensation applied at joint 3 $-4.5°$, at joint 4 $0.5°$.

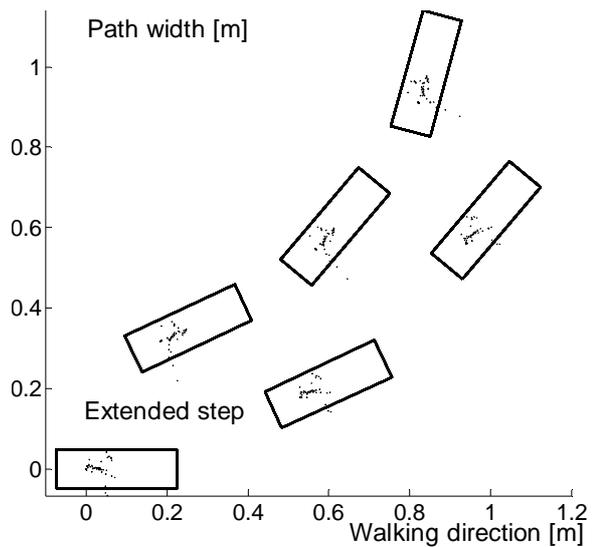

Fig. 44. Mechanism footprints of extended gait in turning left when $\alpha$ is $25°$. No compensation applied.

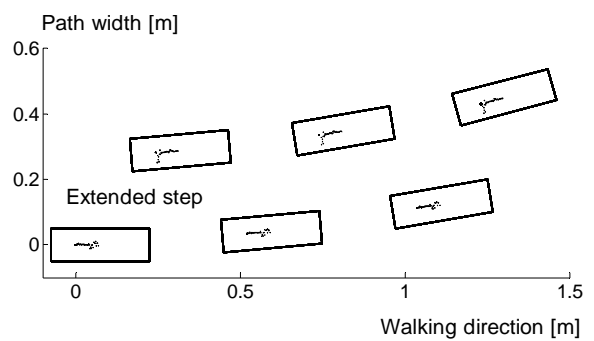

Fig. 45. Mechanism footprints of extended gait in turning left when $\alpha$ is $5°$. No compensation applied.

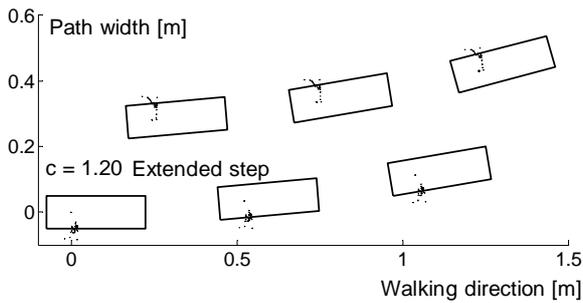

Fig. 46. Mechanism footprints of extended gait in turning left when $\alpha$ is 5° and accelerated motion for $c$ =1.2. No compensation applied.

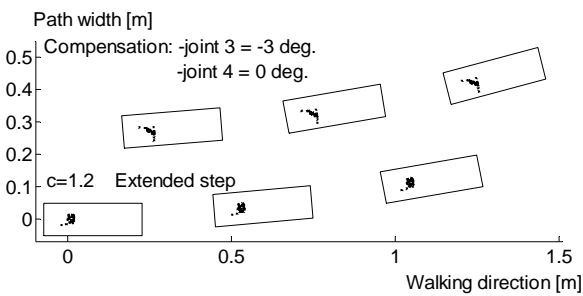

Fig. 47. Mechanism footprints of extended gait in turning left when $\alpha$ is 5° and accelerated motion for $c$ =1.2. Compensation applied at joint 3 –3°, at joint 4 0°.

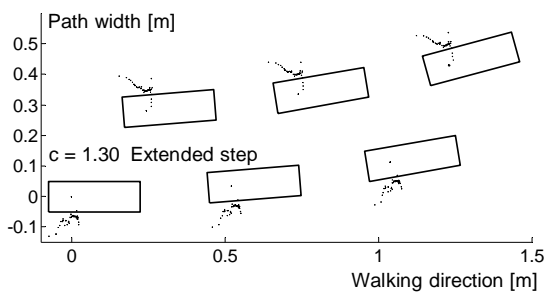

Fig. 48. Mechanism footprints of extended gait in turning left when $\alpha$ is 5° and accelerated motion for $c$ =1.3. No compensation applied.

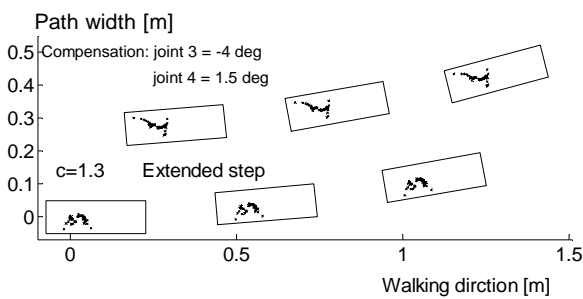

Fig. 49. Mechanism footprints of extended gait in turning left when $\alpha$ is 5° and accelerated motion for $c$ =1.3. Compensation applied at joint 3 –4°, at joint 4 1.5°.

It is again demonstrated that simple inclination in the appropriately selected joint (ankle joint is again used for compensation) can ensure dynamic equilibrium in spite of the nominal gait modification.

Let us now consider a combination of step extension, turning left and faster walking (Figs. 44-49). In Figs. 44, 45, 46 and 48 is presented motion without any compensation. Situation is much worse whem walk is accelerated (Figs. 46 and 48). We applied, again, same compensation strategy. Ankle joint was employed (joints 3 and 4) for compemsation in both, frontal and saggital planes. It is demnstrated that such simple strategy works and in this case

## 5. Conclusions

We demonstrated the applicability of our approach to solving the problem of modification of nominal walk to achieve turning, speed-up and slow-down motion. Deviations of the ZMP position could be to a large extent compensated for simply by incling the body. For the purpose of performing compensation, most appropriate joints were considered and selected.

**Acknowledgement**

This work was funded by Ministry for technology and development of Republic of Serbia under contract MIS 3.04.0019.A/1.